\documentclass[10pt,twocolumn,letterpaper]{article}

\usepackage{cvpr}              % To produce the CAMERA-READY version
\usepackage{times}
\usepackage{epsfig}
\usepackage{graphicx}
\usepackage{amsmath,amsthm}
\usepackage{amssymb}
\usepackage{mathtools}
\usepackage{makecell}
\usepackage{bm,multirow}
\usepackage{multicol}
\usepackage{caption}
\usepackage{subcaption}
\usepackage{adjustbox}
\usepackage{arydshln}

\usepackage{threeparttable}
\usepackage[pagebackref=true,breaklinks=true,colorlinks,bookmarks=false]{hyperref}

\newcommand\blfootnote[1]{%
  \begingroup
  \renewcommand\thefootnote{}\footnote{#1}%
  \addtocounter{footnote}{-1}%
  \endgroup
}
\def\cvprPaperID{10441} % *** Enter the CVPR Paper ID here
\def\confName{CVPR}
\def\confYear{2022}

\begin{document}

%%%%%%%%% TITLE
\title{Exploring Patch-wise Semantic Relation for Contrastive Learning in Image-to-Image Translation Tasks}

\author{Chanyong Jung$^{\ast1}$ \quad\quad\quad Gihyun Kwon$^{\ast1}$ \quad\quad\quad Jong Chul Ye$^{1,2}$\\
Department of Bio and Brain Engineering$^1$, %KAIST  \quad\quad 
Graduate School of AI$^2$, KAIST
\\
\tt\small \{jcy132, cyclomon , jong.ye\}@kaist.ac.kr
}

\maketitle

\blfootnote{$^\ast$co-first authors.}
\blfootnote{This work was supported by Institute of Information \& communications Technology Planning \& Evaluation (IITP) grant funded by the Korea government(MSIT) (No.2019-0-00075, Artificial Intelligence Graduate School Program(KAIST))}
%%%%%%%%% ABSTRACT
\begin{abstract}
%Enhancing the correspondence of input and translation output helps to improve the image translation performance for one-sided translation framework. 
Recently, contrastive learning-based image translation methods have been proposed, 
which contrasts different spatial locations to enhance the spatial correspondence.
However, the methods often ignore the diverse semantic relation within the images. 
To address this,  here we propose a novel semantic relation consistency (SRC) regularization along with the decoupled contrastive learning, which
 utilize the diverse semantics by  focusing on the heterogeneous semantics between the image patches of a single image.
To further improve the performance, we present a hard negative mining by exploiting the semantic relation. 
We verified our method for three tasks: single-modal and multi-modal image translations, and GAN compression task for image translation. Experimental
results confirmed the state-of-art  performance of our method in all the three tasks.

\end{abstract}

%%%%%%%%% BODY TEXT
\section{Introduction}

One of the main objectives of image-to-image translation tasks is to learn a mapping function from a source domain to a target domain so that
it preserves contents while converting the appearance similar to target domain. The cycle consistency loss, which enforces the consistency between the input image and reconstructed image by an inverse mapping of the converted image, is widely used in various frameworks \cite{cycleGAN, stargan_v2, unit, munit}.
However, it requires an additional generator and discriminator to learn the inverse mapping. Also, \cite{cut, HardNegCut, drit} claimed that the cycle-consistency constraint may produce distortion due to its overly restrictive constraint.  
Thus, one-sided image translation methods are suggested to bypass the cycle-consistency constraint by enhancing the correspondence between the input and output in various ways \cite{gcgan, distanceGAN, travelGAN}.

Recently, inspired by the success of contrastive learning, CUT\cite{cut} firstly utilized the contrastive learning to maximize the mutual information between the same location of input and output images. The authors of NEGCUT\cite{HardNegCut} proposed a contrastive learning method using hard negative samples generated  by the negative generator. However, %it relies on the learned distribution by the negative generator, which is not guaranteed to be ‘real’ distribution of negative samples. Also,
 the method requires additional training procedure for the negative generator from random vectors, which may not guarantee
to follow the real negative sample distribution and cause the instability of training.  F-LSeSim\cite{sesim} utilized the patch-wise similarity map for the contrastive learning, but  they ignored the semantic relation between the patches, taking all negative patches as equal negative.

\begin{figure}[!t]
    \includegraphics[width=0.99\linewidth]{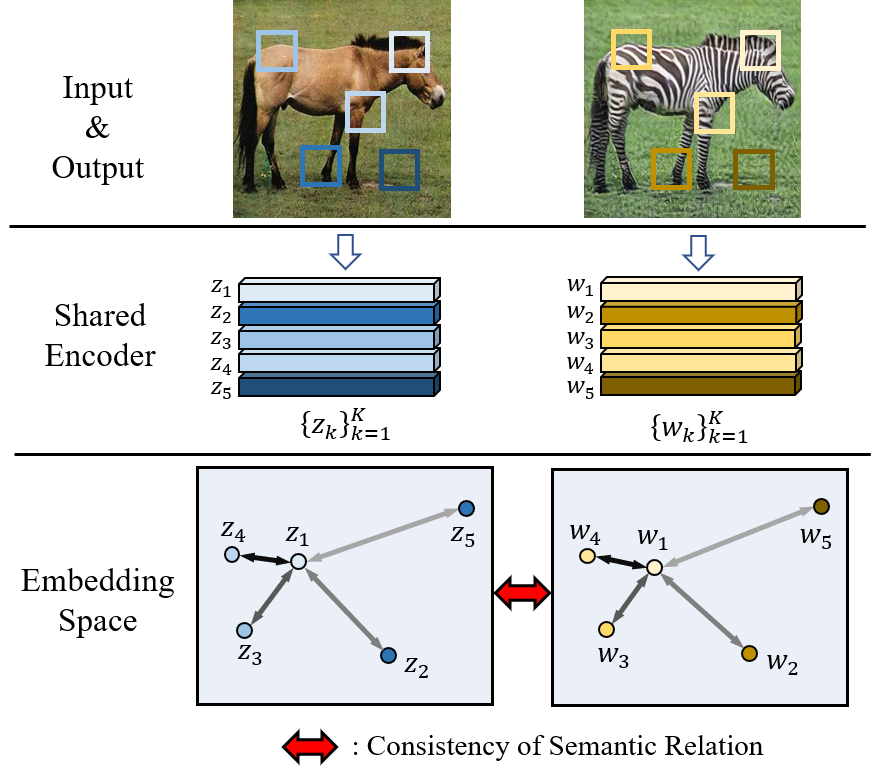}
    \caption{Concept of the proposed method. Decoupled contrastive learning forms the embedding. Consistency regularization of diverse semantic relation is imposed to enhance the correspondence between the input and output.}
    \label{fig:concept}
\end{figure}

In this paper, we propose a novel contrastive learning method to utilize the heterogeneous semantics within an image. 
Specifically, as shown in Fig.~\ref{fig:concept}, the key idea is to impose the consistency regularization on the similarity relation that preserves the spatial semantic relation during the image translation tasks. Specifically, we capture the patch-wise semantic relationship with an image in terms of distributional similarity and  enforce it
to be preserved during the image translation tasks. This semantic regularization prevents from generating image artifacts which violate
semantical relationship.
Furthermore, we propose the hard negative  mining strategy based on the spatially 
varying semantic relations between image patches. This strategy
further improve the performance by avoiding sampling `easy' negative samples which has unrelated semantic information, but more focusing on `hard' negative samples.
%that distance separated patches.
%
%%%Lastly, we further improve the performance by hard negative mining strategy utilizing the relational knowledge. 
%% The curriculum learning technique is applied, controlling the hardness of the negative mining.
%%
%Finally,  the decoupled contrastive learning (DCL) is used instead of InfoNCE criterion
%to prevent the diminishing effect of gradient caused by the semantically unrelated image. 

%The results verifies the proposed method take full advantage from the relative knowledge, 
Experimental results using single- and multi-modal translation task and GAN compression
confirmed that our method produces the state-of-the-art (SOTA) performance 
thanks to its capability of utilizing semantic relationship.

%This paper is composed as follows. We first briefly introduce the related works in Section \ref{sec:related_works}, In Section \ref{sec:method}, we introduced two losses to capture the relative knowledge and impose the consistency regularization. In Section \ref{sec:experiment}, we explain our experimental settings to verify our method in three tasks: single-modal and multi-modal image translation and model compression for image translation. In Section \ref{sec:result}, we present the results and verify our method, which is followed by the conclusion in Section \ref{sec:conclusion}

%-------------------------------------------------------------------------
\section{Related works} \label{sec:related_works}

\subsection{One-sided image translation}

%Inspired by the correspondence 
In order to replace the cycle-consistency, 
many one-sided image translation methods utilize the relational knowledge and correspondency between the input and translated images.
%replacing the cycle-consistency. 
For example, GcGAN\cite{gcgan} utilized the consistency for the geometric transformation of images, and DistanceGAN\cite{distanceGAN} imposed the consistency regularization for the mutual information within a group of images.  The approach in TraVeLGAN\cite{travelGAN} preserved the arithmetic property of embedding vectors.

%As discussed before,
Many contrastive learning methods have been recently suggested to maximize the mutual information between the same location of input and output images
\cite{cut,sesim,HardNegCut}. However, each method has major shortcomings as described before. 
In fact, our method is designed to overcome the limitation of the existing works by exploiting on the relational knowledge
transfer as described below.

\subsection{Relational knowledge transfer}
%Since the representational knowledge is structured \cite{crd}, 
The relational knowledge that  captures the structured interdependencies between the data  is useful for knowledge distillation (KD) tasks \cite{crd, rkd, crcd, sskd}. 
In particular, the student model additionally leverages the inter-sample relation learned by the teacher model, so
the knowledge transfer is proceeded in a  more effective way.

Various methods have been proposed to capture the relational knowledge, such as the angle-wise and distance-wise relation \cite{rkd}, instance-wise correlation within the feature space \cite{cckd}, contrastive relation \cite{sskd}, learned relation for features and gradients by a network \cite{crcd}, etc.

%\cite{rkd} suggested angle-wise distillation loss and distance-wise distillation loss to capture higher-order relational knowledge. \cite{cckd} matched the instance-wise correlation within the feature space. In \cite{sskd}, student model mimics the contrastive relation of teacher model in the embedding space. \cite{crcd} utilized a relation network to learn relational knowledge within the feature space and gradient space. 

%Based on the view for the structured knowledge, 
Accordingly, we are interested in utilizing the semantic relational knowledge to enhance the input and output correspondence for the image translation tasks.
As many previous KD approaches transferred the relational knowledge to enhance the correspondence between student features and teacher features, we match the semantic relational knowledge to enhance the correspondence between input features and output features for image translation tasks.

%In contrast to the existing methods,
Specifically,
our method focused on the heterogeneous semantic property of the image patches,
 which was not considered in the previous works \cite{cut, sesim}. Also,  in contrast to  NEGCUT\cite{HardNegCut},
the negative samples are obtained from the ‘real’ data distribution and a hardness of the negative mining can be controlled by curriculum learning
scheme. 
%Our controllable negative mining strategy enables the curriculum learning, which can further improve the performance. 

%The details for our proposed work is explained in Section~\ref{sec:method}.

\subsection{Contrastive learning}
Contrastive learning method is a framework which obtains useful representation by using the relation between the positive and negative pairs. 
The InfoNCE is widely used for the contrastive learning in many previous works\cite{amdim, moco, simclr, mocov2, cmc}. 

Recently, various approaches are suggested to further improve the performance.
One approach is to sample informative negatives using hard negative mining by von Mises-Fisher distribution \cite{HardNeg}, adversarial training \cite{adco} and learning to rank the samples \cite{siamTriplet}.
%utilize a hard negative mining to sample informative negatives \cite{HardNeg} suggests hard negative mining with von Mises-Fisher distribution. \cite{adco} utilized the adversarial training to obtain hard negative samples. \cite{siamTriplet} selected the negatives by leaning to rank the samples. 
Another approach is to improve the infoNCE loss. For example, FlatNCE \cite{flatclr} is proposed to alleviate the degradation caused by small number of negatives. Decoupled infoNCE loss\cite{dcl} is suggested as the remedy for a negative-positive coupling (NPC) effect, which results the equivalent loss function of FlatNCE.

%Lastly, some works focused on the heterogeneous semantic relation of negative samples. In this approach, some negative samples may have similar semantic information with the anchor data, and others may not. Hence, 
As it is undesirable to treat the negatives equally regardless of their semantic relation,
some works focused on the heterogeneous semantic relation of negative samples. For example,
CO2\cite{co2} proposed the consistency regularization for positive pairs to have same semantic relation with the negatives. PCL\cite{pcl} utilized EM algorithm to encode the semantic structures of a data.

%-------------------------------------------------------------------------
\section{Main Contribution}\label{sec:method}

%We propose a novel method for image translation which utilize the semantic relation within an image. We first capture the similarity relation  of the patches in the embedding space. Then, we impose the consistency regularization for the relations to enhance the spatial correspondence between the input and translation output, using two losses introduced in the following subsections.
%Lastly, we utilize the relative knowledge for the hard negative mining for further improvement. 

\begin{figure}[h!]
    \includegraphics[width=0.98\linewidth]{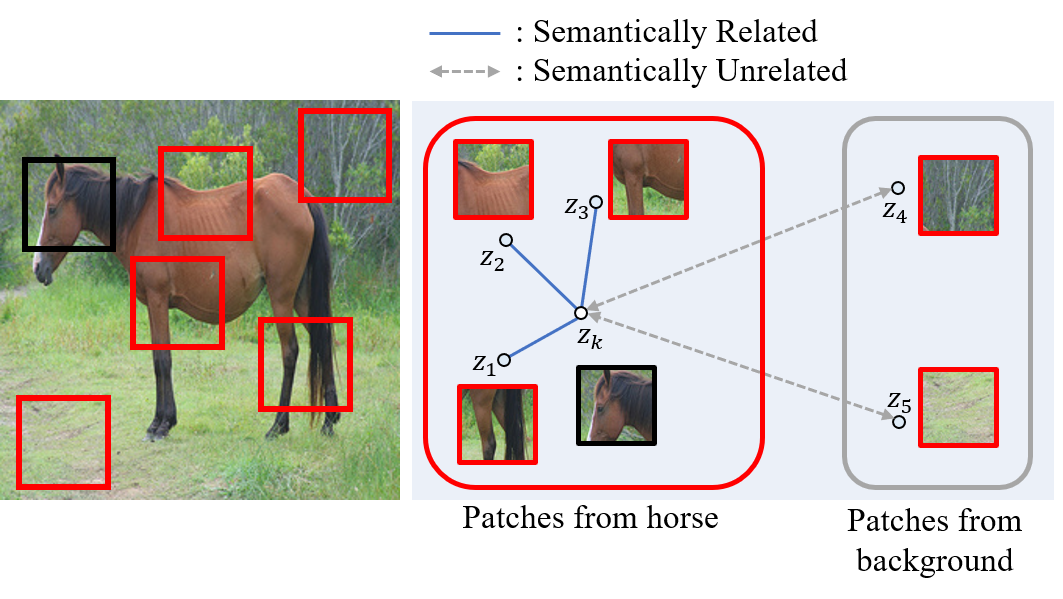}
    \vspace*{-0.3cm}
    \caption{Heterogeneous semantic relation between patches. The black patch is the query and red patches are negatives. Patches from horse are semantically related, but background patches are not.}
    \label{fig:semantic_hetero}
\end{figure}

In natural images, image patches from various locations have heterogeneous semantic relation. For example,
 in Fig. \ref{fig:semantic_hetero}, some patches are from horse, while some others represents the background which are unrelated to the horse. Also, even the patches of the horse have diverse semantic information as they represents each part of the horse such as head and legs. The diverse semantic relation should be considered and preserved for accurate image translation tasks.

In the following subsections, we present our method that utilizes the relational knowledge formed by the patch-wise heterogeneous semantics. 
%Then, we propose a loss for the consistency regularization to enhance the correspondence between the input and output images.

\subsection{Consistency of semantic relation distribution}

%As shown in Fig.\ref{fig:semantic_hetero}, the relational knowledge formed by the patch-wise heterogeneous semantics exists within the image. To enhance the correspondence between the input and output images, which is a main approach for the one-sided translation tasks, the relation should be preserved during the translation procedure. 

\begin{figure}[h!]
	\centering
    \includegraphics[width=0.96\linewidth]{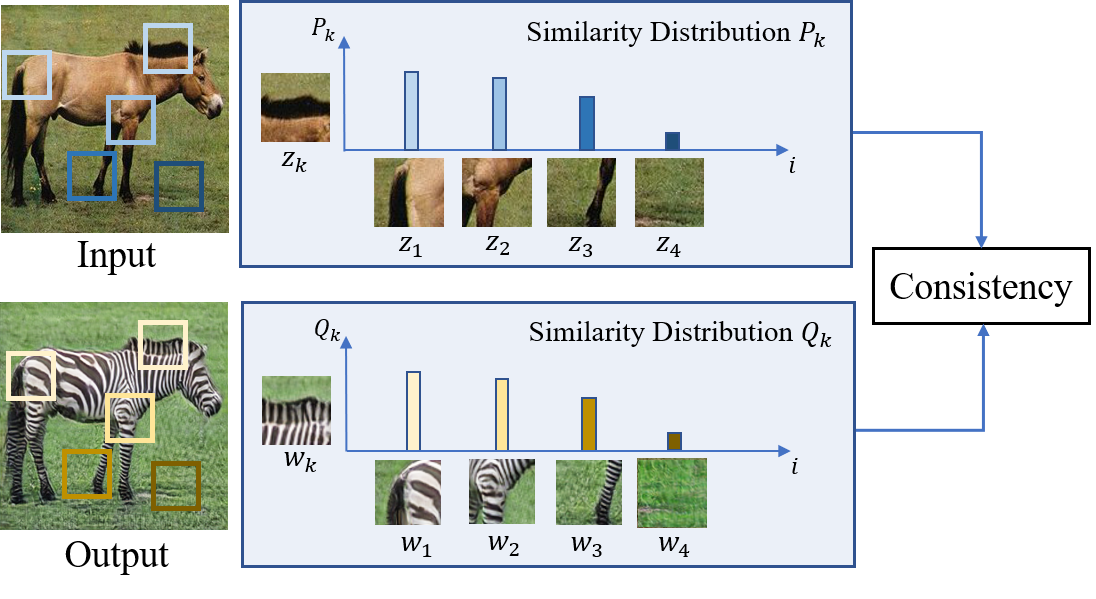}
    \vspace*{-0.2cm}
    \caption{Semantic relation through similarity distribution.}
    \label{fig:sim_consist}
\end{figure}

Let $z_k$ and $w_k$ denote the embedded vectors of image patch $x_k \in \mathcal{X}$ and $y_k \in \mathcal{Y}$, where $\mathcal{X}$ and $\mathcal{Y}$ are input image and translation output image, respectively. The indices $k$ refer to the patch location.

From the perspectives that the contrastive learning is a instance classification \cite{dcl,co2,pcl,i-mix} (i.e. assigning label 1 for positive sample and 0 for negatives), the distribution of similarity can be viewed as a soft label which reveals the structure for semantic relation between the samples \cite{co2, mcl}.
We capture the patch-wise semantic relation as a distribution of similarities, as shown in Fig.~\ref{fig:sim_consist}. 

Accordingly, for  a given  patch $x_k$ of the input image, 
the similarity relation with the negative patch $x_i$ is defined using the soft-max:
%from $i$-th location patch
\begin{equation}
    P_{k}(i) = \frac{\exp{(z_k^\top  z_i)}}{ \sum_{j=1}^K \exp{(z_k^\top  z_j)} }
\end{equation}
where the $z_k$ and $z_i$ are the corresponding embedding vectors.
Then, $P_{k}(i)$ is the distribution to capture the semantic closeness between the $i$-th location and $k$-th location patches within the input image. 
%refers the probability of $i$-th location to be sampled for given $k$-th location patch, which captures the semantic closeness between the patches of the input image.
Similarly, the distribution $Q_{k}(i)$ for the output image similarity relationship is defined as: 
\begin{equation}
    Q_k(i) = \frac{\exp{(w_k^\top  w_i)}}{ \sum_{j=1}^K \exp{(w_k^\top  w_j)} }
\end{equation}
where the $w_i$ and $w_k$ are the embedding vector for the corresponding patches.

To preserve the diverse semantic relation between the patches before and after the image translation,
we impose the consistency regularization on the similarity relation for all $K$ sampled vectors by Jensen-Shannon Divergence (JSD)\cite{co2}:

\begin{equation} \label{eq:L_src_total}
    %L_{sim} = \sum_{k=1}^K L_{sim}^k  \quad \mathrm{where} \quad L_{sim}^k = JSD(q_{z}(~\cdot~| k ) || q_{w}(~\cdot~| k) )
    L_{SRC} = \sum_{k=1}^K JSD(P_{k} || Q_{k} )
\end{equation}

%where 
%\begin{equation} \label{eq:L_sim}
%   L_{sim}^k = JSD(q_{z_k} || q_{w_k} )
%\end{equation}
The minimization of \eqref{eq:L_src_total} therefore enforces the
consistency of the diverse semantic relationship.

%The consistency of the relation captured by the \eqref{eq:L_src_total} imposes the diverse semantic relation to be preserved for the translation procedure. 

\subsection{Hard negative mining by semantic relation}

In this section, we explain how the semantic relation can be utilized for a hard negative mining. % Inspired by the heterogeneous semantics, we use the semantic closeness to sample informative negatives. 
We design the distribution for the hard negative mining and derive the connection with the similarity relation. 
We used decoupled infoNCE loss (DCE)\cite{dcl} for a patch-wise contrastive loss to prevent the negative-positive coupling (NPC) effect which is discussed in detail in the Section~\ref{sec:DCE}.

Specifically, for a given positive pair $(z,w) \sim p_{ZW}$, the hard negative contrastive loss $L_{hDCE}$ using DCE\cite{dcl} with $N$ negatives and temperature parameter $\tau$ is defined as
\begin{align}
    &L_{hDCE}(\gamma, \tau) \notag \\
    &= \mathbb{E}_{(z,w) \sim p_{ZW}} \left[ -\log \frac{ \exp ( w^\top z / \tau ) }
    { N \mathbb{E}_{z^- \sim q_{Z^-}} \left[ \exp ( w^\top z^- / \tau ) \right]  } \right]
\end{align}
where 
%We propose the negative sample mining fo
%r $z \sim p_Z$ 
the negative sampling is modeled by von Mises-Fisher distribution \cite{HardNeg}:
\begin{equation}\label{eq:Pz}
    z^- \sim q_{Z^-}(z^- ; z, \gamma ) = \frac{1}{N_q} \exp{\{\gamma (z^\top z^-)\}} p_Z(z^-)
\end{equation}
where $N_q$ is a normalization constant and $\gamma$ is a  hyperparameter to determine the hardness of the negative samples.

\begin{figure}[!h]
    \includegraphics[width=0.99\linewidth]{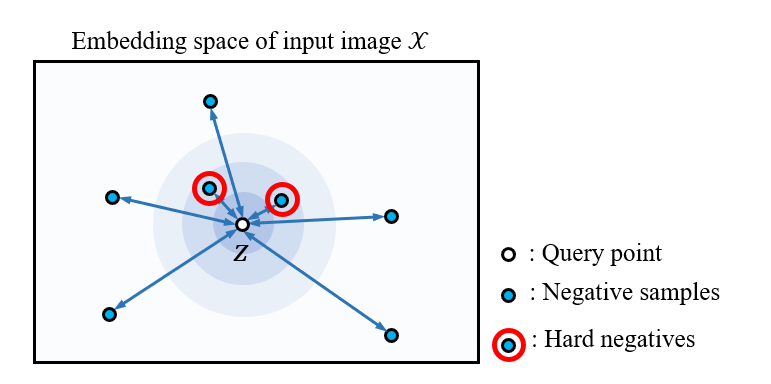}
    \centering
    \caption{Hard negative mining with semantic relation. The color represents the probability to be sampled.}
    \label{fig:HardNeg}
\end{figure}

Accordingly, the designed distribution $q_{Z^-}$ assigns higher probability to be sampled for the negatives $z^-$ which are closer to the $z$, as shown in Fig.~\ref{fig:HardNeg}. By employing this sampling strategy, we can avoid `easy' negative samples that do not contribute the contrastive mechanism significantly.
Instead, by allowing contrastive mechanism between the positive and `hard' negatives, the network
can be trained to have more discriminative power.

In terms of implementation, $\mathbb{E}_{ z^- \sim q_{Z^-} } \left[ \exp ( w^\top z^- / \tau ) \right]$ is approximated with the importance sampling \cite{HardNeg}: 
 
\begin{align*}
    \vspace*{-0.3cm}
    &\mathbb{E}_{z^- \sim q_{Z^-} } \left[  \exp ( w^\top z^-  / \tau) \right]  \notag \\
    &=\mathbb{E}_{z^- \sim p_{Z} } \left[ \exp ( w^\top z^-  / \tau) \cdot \frac{q_{Z^-}(z^-)}{p_Z(z^-)} \right]  \notag \\ 
    &=N_q^{-1}\mathbb{E}_{z^- \sim p_{Z} } \left[  \exp ( w^\top z^-  / \tau) \cdot \exp{\{ \gamma  ( z^\top z^- ) \}} \right]  %\label{eq:HNegContrast}  
\end{align*}
Again, the similarity relation $\exp{\{ \gamma (z^\top z^-)\}}$ becomes the weight of importance sampling for the hard negative mining. 
Finally, curriculum learning is implemented, controlling the hardness of negatives with $\gamma$. Since the semantic relation is unstable at the early of the training, $\gamma$ is initialized as small value. Then, we gradually increase $\gamma$ to sample harder negatives as the training proceeds. 

The explicit control of the hardness for the hard negative mining improved the performance, compared to the NEGCUT\cite{HardNegCut} which used implicit control of the hardness by the negative generator. In the supplementary material, we present the details for the experimental settings, and the effect of $\gamma$ to the performance.

%Differently with [NegCUT] which used the 'learned' distribution by the negative generator, we use 'real' distribution for the negative samples \textcolor{red}{which is obtained from the images of the dataset.}

%Finally, the contrastive loss with hard negatives sampled by the contrast relation is as following, 
%\begin{align}
%    &\mathbf{L}_{contrast}\\
%    &= \mathop{\mathbb{E}}_{x_k,y_k \sim p_{XY}} \left[ -\log \frac{ \exp \{ f(x_k)^T f(y_k) \} }
%    {\frac{1}{K} \sum_{j \neq k } \left[ f(x_j)^T f(y_k) \cdot \exp{\{ \lambda f(x_j)^T f(x_k) \}} \right]  } \right]
%\end{align}

In summary, the total loss considering the patch-wise heterogeneous semantic relation is given by:
\begin{equation}
    L_{semantic}(\gamma, \tau) \notag = \lambda_{SRC} L_{SRC} + \lambda_{hDCE} L_{hDCE}(\gamma, \tau)
\end{equation}
where $\lambda_{SRC}$ and $\lambda_{hDCE}$ are weighting parameters.

%------------------------------------------------------------------------

\begin{figure*}[!h]
    \includegraphics[width=0.99\linewidth]{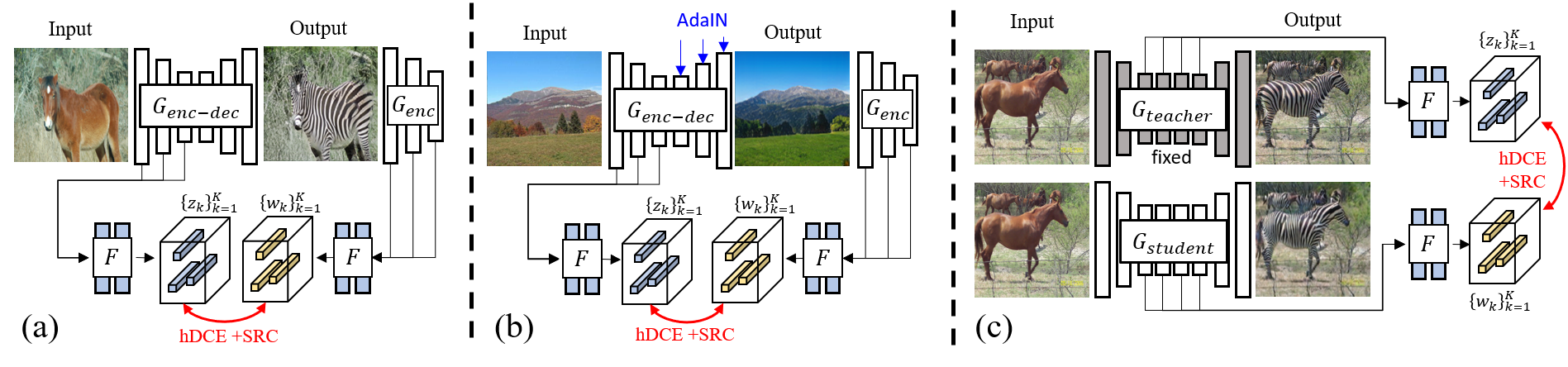}
    \vspace*{-0.2cm}
    \caption{Experiment schematics for (a) single-modal translation, (b) multi-modal translation, and (c) model compression. $F$ consists of sub-networks $F^l$ which is 2-layer MLP for the $l$-th layer. 256 vectors are randomly sampled (i.e $K=256$). }
    \label{fig:exp_setting}
    \vspace*{-0.2cm}
\end{figure*}

\begin{figure*}[!t]
    \includegraphics[width=0.99\linewidth]{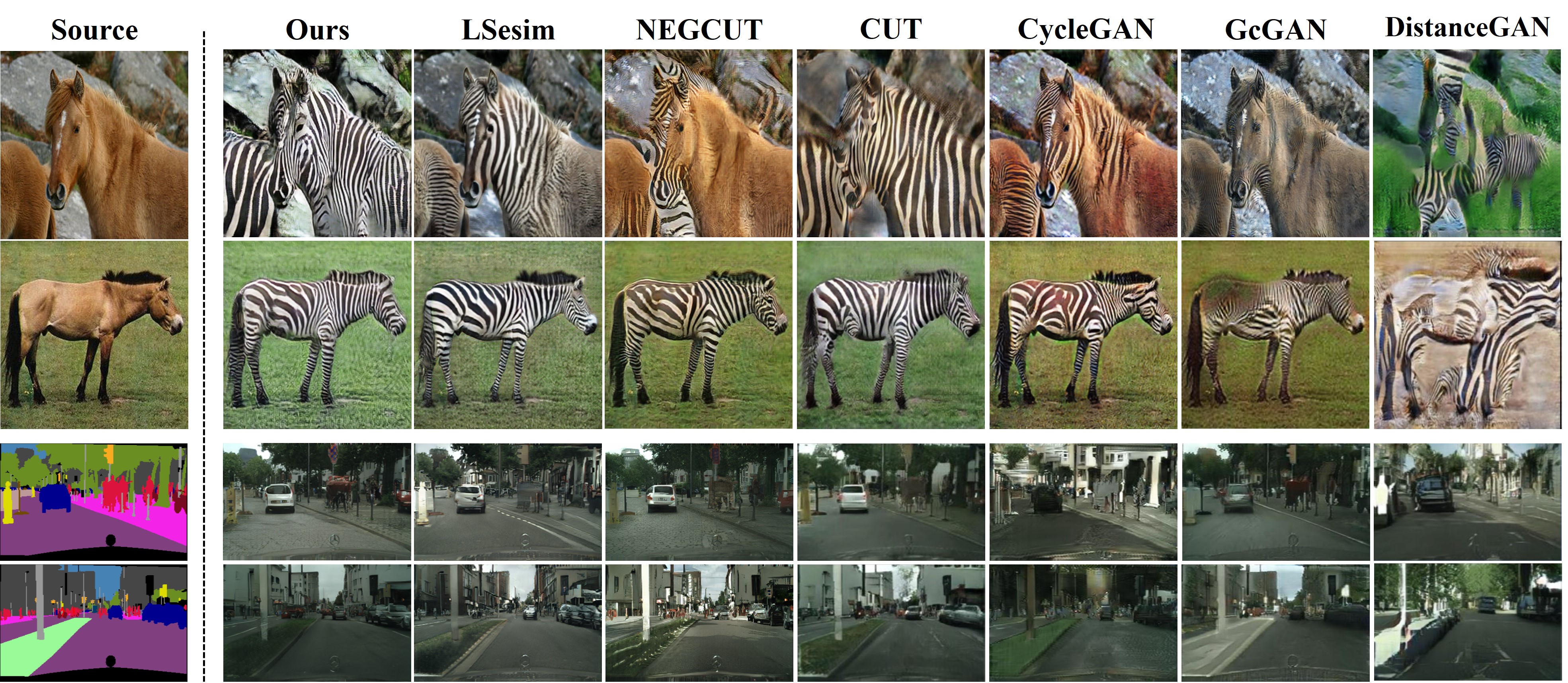}
    \vspace*{-0.2cm}
    \caption{Qualitative comparison on single-modal image translation. Our method shows cleaner and semantically realistic outputs compared with those of baseline methods.}
    \label{fig:result_main}
    \vspace*{-0.3cm}
\end{figure*}

\subsection{Effect of DCE for heterogeneous semantics}\label{sec:DCE}

Decoupled contrastive learning using DCE\cite{dcl} is proposed to remove the negative-positive coupling (NPC) effect. Specifically, the NPC effect refers the diminishing gradient of the InfoNCE by the easy negatives and positive samples, hindering the update by the other informative samples. 

Likewise, for the InfoNCE based image translation methods \cite{cut,HardNegCut,sesim}, the NPC effect is problematic due to the heterogeneous semantic relation between the image patches. 
Since we randomly sample the negatives, semantically unrelated image patches can be included as the negatives.
 For example, $z_4$ and $z_5$ in the Fig.~\ref{fig:semantic_hetero} are
 unrelated easy negatives which cause the NPC effect.
To prevent the NPC effect, as discussed before, decoupled InfoNCE (DCE) loss should be employed in this paper.
This section is dedicated to provide explicit explanation why NPC effect in the classical InfoNCE loss is detrimental for image translation, and why DCE is a better alternative for image translation.

Specifically, for a positive pair $(z_k, w_k)$, the DCE loss is defined as:
 \begin{align}\label{eq:DCE}
    L_{\mathrm{DCE}} = - \log \frac{\exp{(z_k^\top w_k / \tau)}}{\sum_{j \neq k} \exp{(z_j^\top w_k / \tau)} } 
\end{align}
which removes the positive pair term in the denominator from the InfoNCE loss:
\begin{align}\label{eq:Info}
    &L_{\mathrm{InfoNCE}} \notag \\
    & = - \log \frac{\exp{(z_k^\top w_k / \tau)}}{\exp{(z_k^\top w_k / \tau)} + \sum_{j \neq k} \exp{(z_j^\top w_k / \tau)} } 
\end{align}

As discussed in \cite{dcl}, the gradient from the loss function $\nabla_{w_k} L$ for the contrastive loss $L$ in \eqref{eq:DCE} and \eqref{eq:Info} is given as: 
%Then, the following proposition explicitly
%shows the gradients of the two losses: % shows the NPC effect and the superiority of DCE for heterogeneous semantic features, in formal form. 

% The easy negative $z_j$ has unrelated semantic information with $z_k$ which results a small value of $\exp{(z_j^\top z_k / \tau)}$. Hence, the easy negative contributes to the diminishing gradients from the contrastive learning, which is known as NPC effect.  

%Thereby, the gradients from the loss do not diminshed by the easy negatives, preventing from the NPC affect by the semantically unrelated image patches.

\begin{align*}
    &\nabla_{w_k} L = -\frac{\alpha}{\tau} \left \{ z_k - \sum_{j \neq k} \frac{ \exp{(z_j^\top w_k / \tau)} }{ \sum_{m \neq k} \exp(z_m^\top w_k / \tau )} \cdot z_j \right \} 
\end{align*}
where
\begin{align}\label{eq:alpha}
\alpha \coloneqq
\begin{cases}
q_{NPC} & \text{ if } L=L_{\mathrm{InfoNCE}} \\ 
1 & \text{ if } L=L_{\mathrm{DCE}}
\end{cases}
\end{align}
and
\begin{equation}\label{eq:qNPC}
q_{NPC} \simeq 1- \frac{ \exp (z_k^\top w_k / \tau) }{ \exp{(z_k^\top w_k / \tau)} + \sum_{j \neq k} \exp{(z_j^\top z_k / \tau)} } 
\end{equation}

If $z_j$ for $j\neq k$ has unrelated semantic information with $z_k$, $z_j$ will have a small value of $\exp{(z_j^\top z_k / \tau)}$. 
Hence, the denominator in \eqref{eq:qNPC} is dominated by $ \exp{(z_k^\top w_k / \tau)}$, decreasing the value of 
  $q_{NPC}$ and $\alpha$ through \eqref{eq:alpha}, which again results the diminishing gradient for InfoNCE loss. 
However, in case of DCE, the gradients are not related to the $q_{NPC}$ as $\alpha=1$, which prevents the gradient vanishing by the easy negatives. 
Therefore, DCE loss is employed in our method to prevent the NPC effect caused by the semantically unrelated negative samples.

%The derivation for the proposition \ref{prop_npc} is explained in detail in the Supplemantery Material. 

%\begin{align}
%    &L_{\mathrm{DCNCE}} \notag \\
%    & = - \log \frac{\exp{(z_k \cdot w_k / \tau)}}{\sum_{j \neq k} \exp{(z_j \cdot w_k / \tau)} }
%\end{align}
% with the gradients as, 
%\begin{align}
%    &\nabla_{z_k} L_{\mathrm{DCNCE}} = -\frac{w_k}{\tau} \label{eq:grad_zk} \\
%    &\nabla_{z_j} L_{\mathrm{DCNCE}} = -\frac{ \exp{(z_j \cdot w_k / \tau )} }{ \sum_{i \neq k} \exp{(z_i \cdot w_k / \tau)} } \cdot \frac{w_k}{\tau}  \label{eq:grad_zj} \\
%    &\nabla_{w_k} L_{\mathrm{DCNCE}} \notag \\
%    & \quad = -\frac{1}{\tau} \left \{ z_k - \sum_{j \neq k} \frac{ \exp{(z_j \cdot w_k / \tau)} }{ \sum_{i \neq k} \exp(z_i \cdot w_k / \tau )} \cdot z_j \right \} \label{eq:grad_wk}
%\end{align}

\section{Experiments}\label{sec:experiment}

To show the versatility of our model, we apply our proposed method to various frameworks for image-to-image translation (I2I) tasks, which
 includes single-modal and multi-modal image translation, and I2I GAN model compression. The brief introduction of our experiment is in Fig.~\ref{fig:exp_setting}. 
Additionally, we provide more results for single-image translation task (i.e. \textit{painting}$\rightarrow$\textit{photo}) in the Supplementary Material. 

% \subsection{Image Translation}
\subsection{Single modal image translation}

First, we evaluate our method on unpaired image translation task which is designed to translate  input images to a single output  target domain.

% \begin{table}[!t]
\begin{threeparttable}[!b]
\begin{center}
% \begin{adjustbox}{width=0.7\textwidth}
% \begin{tabular*}{0.7\textwidth}{@{\extracolsep{\fill}}c|cccc|cccc}	

\resizebox{0.48\textwidth}{!}{

\begin{tabular}{@{\extracolsep{5pt}}cccccc@{}}
% \toprule
% \cline{2-11}
\hline
% 	\multirow{2}{*}{CelebA-HQ}& \multicolumn{4}{c|}{CelebA-HQ} & \multicolumn{4}{c}{AFHQ}\\ \cline{2-5}\cline{6-9}

\multirow{2}{*}{\textbf{Methods}} & \textbf{H$\rightarrow$Z} &\multicolumn{4}{c}{\textbf{Cityscapes}}\\ %& \textbf{Cat$\rightarrow$Dog}\\
% \cline{2-11}
% \cmidrule(lr){2-2}
\cline{2-2} 
\cline{3-6} 
%\cline{7-7}
% \midrule
 & FID$\downarrow$&mAP$\uparrow$&pAcc$\uparrow$&cAcc$\uparrow$&FID$\downarrow$\\ %&FID$\downarrow$ \\

\hline
CycleGAN\cite{cycleGAN} &77.2&20.4&55.9&25.4&76.3\\ %&85.9\\
MUNIT\cite{munit} &133.8&16.9&56.5&22.5&91.4\\ %&104.4\\
DRIT\cite{drit} & 140.0&17.0&58.7&22.2&155.3\\ %&123.4\\
\hdashline
Distance\cite{distanceGAN}&72.0&8.4&42.2&12.6&81.8\\ %&155.3\\
GcGAN\cite{gcgan} &86.7&21.2&63.2&26.6&105.2\\ %&96.6\\
CUT\cite{cut}&45.5&24.7&68.8&30.7&56.4\\ %&76.2\\
\hdashline
NEGCUT\cite{HardNegCut} &39.6&27.6&71.4&35.0&48.5\\ %&55.9 \\
LSeSIM$*$\cite{sesim}& 38.0& -- & 73.2& -- &49.7\\ %& \\
OURS& \textbf{34.4} & \textbf{29.0} & \textbf{73.5} & \textbf{35.6} &\textbf{46.4}\\ %&\textbf{62.0} \\

\hline

\end{tabular}
}
% \end{adjustbox}
\end{center}
    \vspace*{-.1cm}
\begin{tablenotes}
\item [$*$] \footnotesize{LSeSim did not report mAP\&cAcc}
\end{tablenotes}
\vspace*{-.3cm}
\caption{Quantitative comparison of single-modal image translation. H$\rightarrow$Z refers to Horse$\rightarrow$Zebra 
dataset. Our method outperforms baseline models.}
\vspace*{0.3cm}
% \vspace*{-.3cm}
\label{table:unpaired}
\end{threeparttable}
% \end{table}

\begin{figure*}[!h]
    \includegraphics[width=0.99\linewidth]{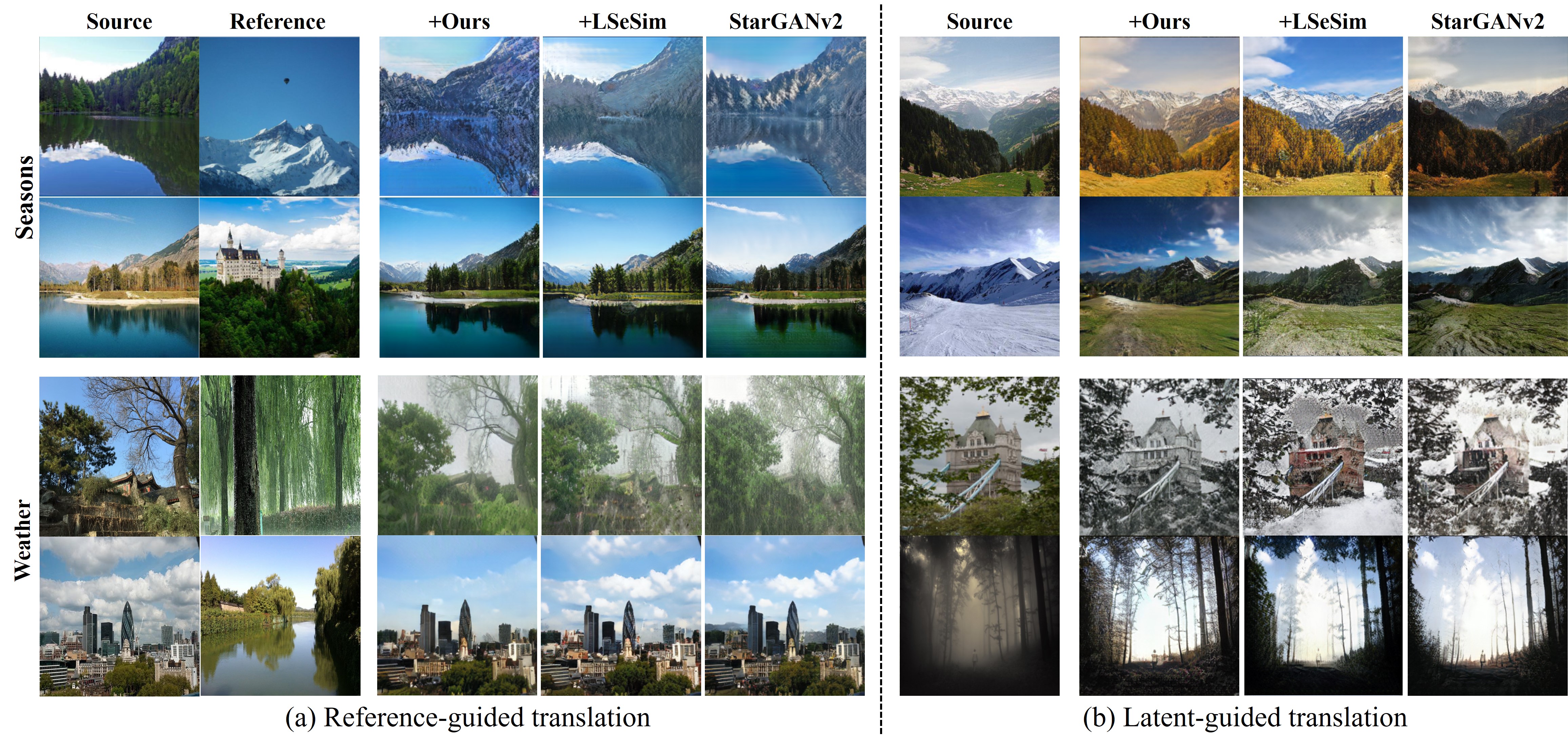}
    \vspace*{-.3cm}
    \caption{Qualitative comparison on multi-modal image translation. 
    Our method can translate source images to arbitrary domains with diverse styles. (a) Reference-guided translation: source images are translated to target domains with reflecting reference style. (b) Latent-guided translation: each method translate source image to target domain with random styles. The rows from the top show translated outputs in the following order: summer2autumn, winter2spring, cloudy2snowy, and foggy2sunny. }
    \label{fig:result_star}
\end{figure*}

\noindent\textbf{Experiment details:} Our implementation of unpaired translation model is based on source code of recent image translation model of CUT\footnote{https://github.com/taesungp/contrastive-unpaired-translation}. %Instead of using basic infoNCE loss of CUT, 
%We trained the models with our proposed methods of decoupled infoNCE with hard negative mining (hDCE) and spatial relation consistency loss(SRC).
% Since our method concentrates on the spatial relation, 
 We evaluate our method on datasets of horse$\rightarrow$zebra and cityscapes label$\rightarrow$images. In the framework of CUT\cite{cut}, we replace the PatchNCE loss with our proposed loss ($L_{semantic}$). Detailed settings are provided in our Supplementary Materials.

\noindent\textbf{Evaluation metrics:} For evaluation, we calculate the quantitative metrics used in the previous works. For the models trained on horse$\rightarrow$zebra dataset, we use Frechet Inception Distance (FID) \cite{fid} to evaluate the translated image quality. In the case of cityscapes (label$\rightarrow$image), we additionally measure the correspondence between the segmented maps of the outputs and their ground truth maps. Specifically, we use pre-trained model of DRN \cite{drn} for segmentation, and calculate mean average precision recall (mAP), pixel-wise accuracy (pixAcc), and average class accuracy (classAcc) \cite{cut}.

\noindent\textbf{Results:} Table~\ref{table:unpaired} shows the quantitative comparison between the proposed method and the previous works. In all of the datasets and metrics, our method outperforms the baseline methods. Specifically, our method outperformed the existing models with large margin when compared with one-sided image translation methods such as DistanceGAN \cite{distanceGAN} and GcGAN \cite{gcgan} and two-sided models such as CycleGAN \cite{cycleGAN}, MUNIT \cite{munit}, and DRIT \cite{drit}. Comparing to recent methods which use contrastive learning (CUT \cite{cut}, NEGCUT \cite{HardNegCut}, LSeSim \cite{sesim}), we also achieved better performance in overall metrics.

For qualitative comparison, we show the results in Fig.~\ref{fig:result_main}. When we compare the results with those of the baselines, in the case of horse$\rightarrow$zebra, the images generated by our method fully reflect the semantic texture of the target domain while the spatial structure of the input source is well preserved. In the experiments on cityscapes, the baselines often produce class mismatch between the input labels and the outputs. On the other hand, our  model generates realistic images with right correspondence to the labels.

%-------------------------------------------------------------------------

\subsection{Multi-modal image translation}

\begin{table}[!h]
% \begin{threeparttable}[!t]
\begin{center}
% \begin{adjustbox}{width=0.7\textwidth}
% \begin{tabular*}{0.7\textwidth}{@{\extracolsep{\fill}}c|cccc|cccc}	

\resizebox{0.48\textwidth}{!}{

\begin{tabular}{@{\extracolsep{5pt}}cccccc@{}}
% \toprule
% \cline{2-11}
\hline
% 	\multirow{2}{*}{CelebA-HQ}& \multicolumn{4}{c|}{CelebA-HQ} & \multicolumn{4}{c}{AFHQ}\\ \cline{2-5}\cline{6-9}

% \multirow{3}{*}{\textbf{Dataset}}&\multirow{3}{*}{\textbf{Methods}} & \multicolumn{4}{c}{\textbf{Seasons}} &\multicolumn{4}{c}{\textbf{Weather}}\\
% \cline{2-5} 
% \cline{6-9} 
% \cline{7-7}
 \multirow{2}{*}{\textbf{Dataset}}&\multirow{2}{*}{\textbf{Method}}  & \multicolumn{2}{c}{\textbf{Latent}} &\multicolumn{2}{c}{\textbf{Reference}}\\
% \cline{2-11}
% \cmidrule(lr){2-2}
\cline{3-4} 
\cline{5-6}
% \cline{6-7} 
% \cline{8-9}
% \midrule
 && FID$\downarrow$&LPIPS$\uparrow$& FID$\downarrow$&LPIPS$\uparrow$ \\

\hline
\multirow{3}{*}{\textbf{Seasons}} &StarGANv2\cite{stargan_v2} &63.06&	0.413&	61.19&	0.346\\
&+LSeSim\cite{sesim} &61.50&0.378&60.40&0.302\\
\cdashline{2-6}
&+Ours & \textbf{54.70}& \textbf{0.496}&\textbf{54.23}&\textbf{0.365}\\
\hline
\multirow{3}{*}{\textbf{Weather}} &StarGANv2\cite{stargan_v2} &62.45&	0.415&	64.20&	0.342\\
&+LSeSim\cite{sesim} &60.07&	0.335	&62.17	&0.286\\
\cdashline{2-6}
&+Ours & \textbf{54.02}&	\textbf{0.470}&	\textbf{56.91}&	\textbf{0.362}\\
\hline

\end{tabular}
}
% \end{adjustbox}
\end{center}
\vspace*{-.3cm}
\caption{Quantitative comparison of multi-modal image translation.}

\vspace*{-.3cm}
\label{table:diverse}
% \end{threeparttable}
\end{table}

\begin{figure*}[!t]
    \includegraphics[width=0.99\linewidth]{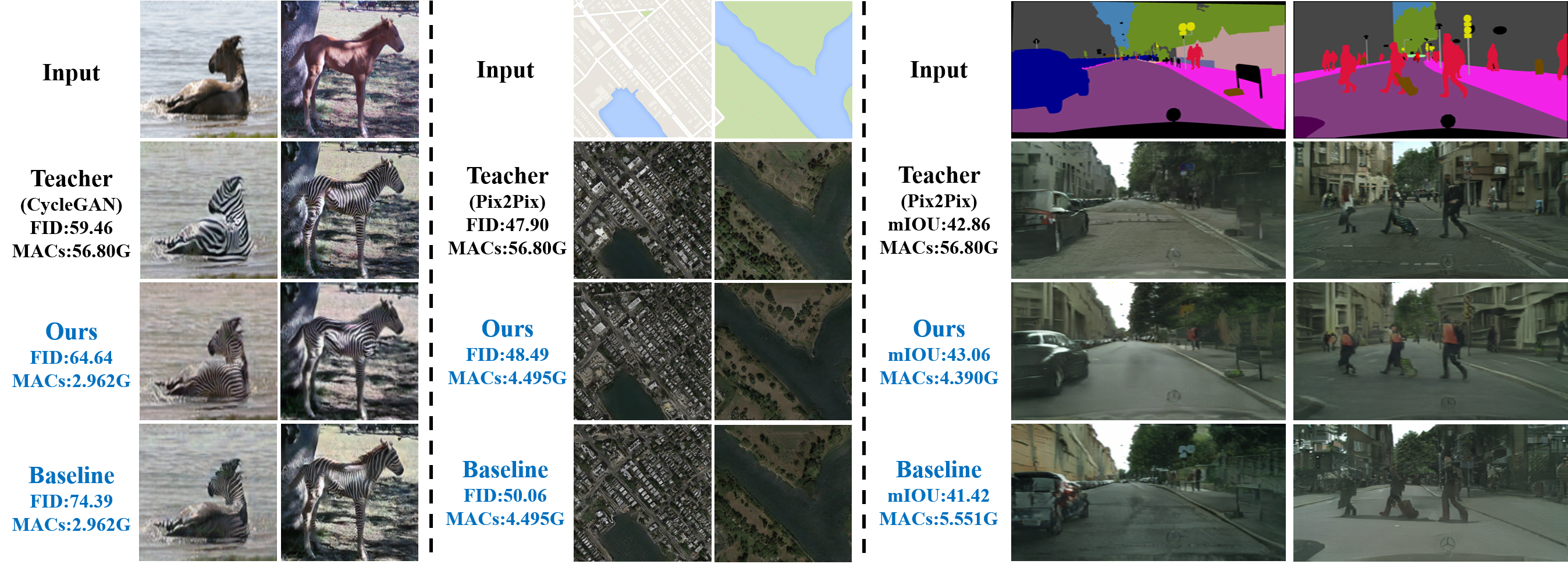}
    \vspace*{-0.2cm}
    \caption{Qualitative comparison on model compression. (left) horse$\rightarrow$zebra, (center) map$\rightarrow$satellite aerial, and (right) cityscapes label$\rightarrow$image. Our method shows high performances with small network size.}
    \label{fig:result_comp}
    \vspace*{-0.4cm}
\end{figure*}

For further evaluation, we apply our method to multi-modal image translation model which is a framework for translating input to diverse outputs with multiple domains. 

\noindent\textbf{Experiment details:} Our method is implemented using the official source code\footnote{https://github.com/clovaai/stargan-v2} of the state-of-the-art diverse translation model StarGANv2\cite{stargan_v2}. Specifically,  on top of the loss functions of basic StarGANv2, we add our proposed loss ($L_{semantic}$). Similar to single-modal translation, we calculate losses by sampling the embedded vectors from the input and output features (Fig.~\ref{fig:exp_setting}(b)). For the comparison with LSesim\cite{sesim}, we implemented the StarGANv2 model with LSeSim following the official source code \footnote{https://github.com/lyndonzheng/F-LSeSim}.

\noindent\textbf{Dataset:} We choose two large-scale multi-domain datasets : {\it{Seasons}}\cite{seasons} and {\it{Weather}}\cite{weather}. {\it{Seasons}} dataset consists of 4 domains (spring, summer, autumn, winter), each domain has 1208,1322,1460,and 1055 images. {\it{Weather}} dataset consists of 5 domains (sunny, cloudy, rainy, foggy, snowy) of which each domain has 4,000 images.
As validation sets, we choose 100 images per each domain for {\it{Seasons}} dataset, and 400 images per each domain for {\it{Weather}} dataset.

\noindent\textbf{Evaluation metrics:} 
We measure the image quality with FID and generation diversity through LPIPS \cite{lpips}. We follow the evaluation scenarios of StarGANv2: 1) latent-based image translation, which converts
the style of input image to a random style, and 2) reference-based image translation, in
which we convert the styles of inputs to those of the reference
images. To calculate metrics, we first generate 10 different outputs per single input. We calculate LPIPS distance for 40 samples, and the average score is obtained by repeating the process for all generated images. We also calculate FID between generated target domain outputs and the training images of the corresponding domain. We report the averaged FID score for all domain cases.

\noindent\textbf{Results:}
We show the quantitative comparison results in Table~\ref{table:diverse}. Our method produced superior performance for all metrics. More specifically, the model trained with SeSim has improved performance in image quality with lower FID scores, but it shows weakness in terms of diversity with decreased LPIPS scores. However, when our method is used, both of image quality and diversity are improved in both datasets with a large margin.

Fig.~\ref{fig:result_star} shows qualitative results on two different datasets. We compared the results of images translated into various domains. In case of the baseline models, the structure correspondence between input and output is not maintained, and 
some artifacts are included in the output as the information of target domain is not fully reflected. 
%often the information of the target domain was not successfully reflected with additional artifacts. 
On the other hand, the result of our method 
better preserved the spatial information of input source, showing the successful translation with diverse styles.

\subsection{Compression of image translation model}

\begin{table}[!t]
% \begin{threeparttable}[!t]
\begin{center}
% \begin{adjustbox}{width=0.7\textwidth}
% \begin{tabular*}{0.7\textwidth}{@{\extracolsep{\fill}}c|cccc|cccc}	

\resizebox{0.48\textwidth}{!}{

\begin{tabular}{@{\extracolsep{5pt}}cccccc@{}}
% \toprule
% \cline{2-11}
\hline
% 	\multirow{2}{*}{CelebA-HQ}& \multicolumn{4}{c|}{CelebA-HQ} & \multicolumn{4}{c}{AFHQ}\\ \cline{2-5}\cline{6-9}

% \multirow{3}{*}{\textbf{Dataset}}&\multirow{3}{*}{\textbf{Methods}} & \multicolumn{4}{c}{\textbf{Seasons}} &\multicolumn{4}{c}{\textbf{Weather}}\\
% \cline{2-5} 
% \cline{6-9} 
% \cline{7-7}
\multirow{2}{*}{\textbf{Dataset}}&\multirow{2}{*}{\textbf{Model}}&\multicolumn{4}{c}{\textbf{Metrics}}\\
\cline{3-6}
  &   & {{\#Param$\downarrow$}} &{{MACs$\downarrow$}}&{{FID$\downarrow$}}&{{mIOU}$\uparrow$}\\

% \cmidrule(lr){2-2}
\hline
\multirow{3}{*}{\textbf{H$\rightarrow$Z}} &Teacher &11.38M&	56.80&	59.46&	- \\
&Baseline &0.412M&	2.962	&74.39	&-\\
% \cdashline{2-6}
&+Ours & 0.412M&	2.962&	\textbf{64.64}&	-\\
\hline
\multirow{3}{*}{\textbf{M$\rightarrow$S}} &Teacher &11.38M&	56.80&	47.90&	- \\
&Baseline &0.667M&	4.495&	50.06	&-\\
% \cdashline{2-6}
&+Ours & 0.667M&	4.495&	\textbf{48.49}&	-\\
\hline
\multirow{3}{*}{\textbf{city}} &Teacher &11.38M&	56.80&60.38&	42.86 \\
&Baseline &0.730M&	5.551	&85.24	& 41.42\\
% \cdashline{2-6}
&+Ours & \textbf{0.685M}&	\textbf{4.390} &	\textbf{72.41}&	\textbf{43.06}\\
\hline
\end{tabular}
}
% \end{adjustbox}
\end{center}
\vspace*{-.3cm}
\caption{Quantitative comparison of model compression. H$\rightarrow$Z refers to Horse$\rightarrow$Zebra 
dataset, M$\rightarrow$S refers to Map$\rightarrow$Satellite 
dataset, and city refers to cityscape label$\rightarrow$image dataset. Our method outperforms baseline method.}

\vspace*{-.3cm}
\label{table:compression}
% \end{threeparttable}
\end{table}

In recent model distillation methods, the delivery of the relational knowledge learned by the teacher model improves the student model, providing the additional information. 
We apply our method for a GAN compression framework, transferring the patch-wise semantic relational knowledge from the teacher to student. We demonstrate the effectiveness of our method for the GAN compression.

%Since our method focuses on maintaining the spatial relation between input-output sides, we apply our method for GAN compression.

\noindent\textbf{Experiment details:} For the experiment, we adopt Fast GAN Compression \cite{Li_2020_CVPR}, a SOTA compression method for image translation models. The method consists of two steps. In step 1, we first train the student network (a.k.a super-net) by distilling the feature of pre-trained teacher generator to student generator. In this step, the student model is a once-for-all network which supports different channel numbers. In step 2, using evolution search, we find the optimal channel number configurations of the student based on the target metric (e.g FID, mIoU). Experimental details are in the Supplementary Material.

We include our proposed method in step 1 of the model distillation part. Specifically, in Fig.~\ref{fig:exp_setting}(c), we introduce header net $F$ and matched the semantic relation between the embedded features of the teacher and the student. The proposed loss ($L_{semantic}$) is used along with the loss functions of the baseline framework of Fast GAN Compression\cite{Li_2020_CVPR}. The details of experimental setting are in the Supplementary Material. For implementation, we referenced the original source code of GAN compression-fast version\footnote{https://github.com/mit-han-lab/gan-compression}.

\noindent\textbf{Dataset and models:} To evaluate the performance, we conduct experiments on 3 different settings : compression of CycleGAN trained on Horse$\rightarrow$Zebra, Pix2Pix trained on Map$\rightarrow$Satellite aerial, Pix2Pix trained on Cityscapes label$\rightarrow$images. We retrained the baseline models using the official code for fair comparison.

\noindent\textbf{Evaluation metrics:} 
To compare the generation quality of the compressed model, we measure FID values. In the case of Cityscape, mIoU between output and GT is additionally measured. We also compare the compression performance with calculating multiply-accumulate operations (MACs) and the number of model parameters (\#Param).

\noindent\textbf{Results:} Table \ref{table:compression} shows the quantitative performance of model compression. The results show that our model has better FID and mIoU scores while having similar or less MACs and \#Param compared to the baseline student model. Particularly, in case of cityscapes dataset, we obtain better mIoU than teacher model. This suggests that our proposed method improves the student model, and additionally delivers the relational knowledge of teacher.

Fig.~\ref{fig:result_comp} is provided for qualitative comparison. Although our compressed model size is much smaller than the teacher model, the quality of the generated images is not worsened. Comparing with the baseline outputs, we can see that the output images have clear boundaries and better target domain textures.

\subsection{Ablations studies}
We compare the quantitative performance of models trained on different settings using Horse$\rightarrow$Zebra and cityscapes datasets for single-modal image translation task.

Specifically, we progressively add the components of our method, and observe the corresponding performance. 
More specifically, we first start from the basic framework using infoNCE, which is identical to CUT \cite{cut}. Then we compare the infoNCE with DCE, sequentially including semantic relation consistency (SRC) and hard negative mining (Hneg).

The results in Table \ref{table:ablation} shows the  meaningful improvement by our method. 
Specifically, when each component is added to DCE, we observe the improvement in both datasets: Horse$\rightarrow$Zebra and cityscapes. When all the components are added, the results showed the best performance.

In addition, in case of InfoNCE loss, the SRC loss and the hard negative mining contributes to the improvement. However, compared with our best model with DCE, the overall results with the InfoNCE loss show worse performance in all metrics. 

% In addition, we conduct the experiments to check the effect of the number of negative samples. As shown in Table \ref{table:ablation}, the results are not dramatically affected by the number of samples, but we observed slight
% performance drop  when different number of negative samples are used.

\begin{figure}[!t]
    \includegraphics[width=0.99\linewidth]{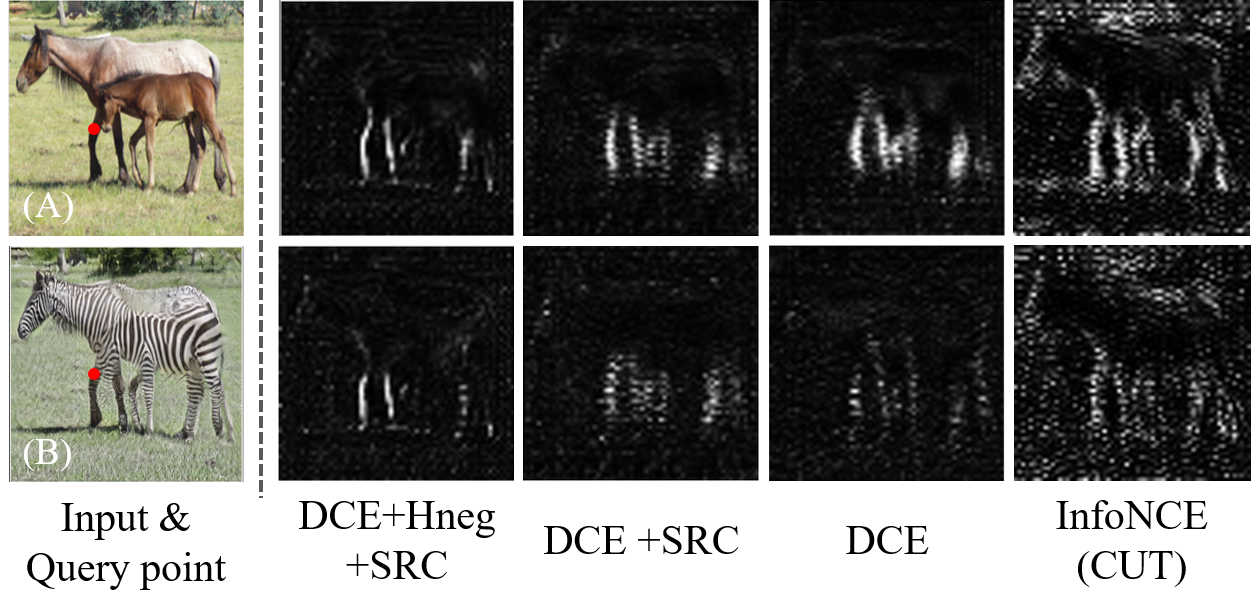}
    \vspace*{-0.3cm}
    \caption{Similarity maps of models trained on horse$\rightarrow$zebra datasets. Red dot at the front leg is the query point.}
    \label{fig:result_ablation}
\end{figure}

To further show the effect of proposed methods, we show the similarity relation in Fig.~\ref{fig:result_ablation}. The learned similarity between the query point and the other locations is calculated and mapped for input image (A) and translation output (B). In case of InfoNCE and basic DCE, high similarity is shown at the points not related to the query (especially severe for infoNCE), and (A) and (B) show different similarity patterns. When SRC is used, the similarity aspect of (A) and (B) become similar. When the hard negative mining is further applied, only the points that are closely related to the query point are embedded with high similarity. It shows that we achieve our goals: consistency of patch-wise semantic relation between (A) and (B) and learning on hard negative samples. 

More ablation studies and additional results on our three different tasks are in Supplementary Materials.

\begin{table}[!t]
% \begin{threeparttable}[!t]
\begin{center}
% \begin{adjustbox}{width=0.7\textwidth}
% \begin{tabular*}{0.7\textwidth}{@{\extracolsep{\fill}}c|cccc|cccc}	

\resizebox{0.48\textwidth}{!}{

\begin{tabular}{@{\extracolsep{5pt}}ccccccccc@{}}
% \toprule
% \cline{2-11}
\hline
% 	\multirow{2}{*}{CelebA-HQ}& \multicolumn{4}{c|}{CelebA-HQ} & \multicolumn{4}{c}{AFHQ}\\ \cline{2-5}\cline{6-9}

\multicolumn{4}{c}{\textbf{Settings}} & \textbf{H$\rightarrow$Z} &\multicolumn{4}{c}{\textbf{Cityscapes}}\\ %& \textbf{Cat$\rightarrow$Dog}\\
% \cline{2-11}
% \cmidrule(lr){2-2}
\cline{1-4} 
\cline{5-5} 
\cline{6-9} 
%\cline{7-7}
% \midrule
 Info&\multirow{2}{*}{DCE} &SRC&Hard Neg& \multirow{2}{*}{FID$\downarrow$}&\multirow{2}{*}{mAP$\uparrow$}& \multirow{2}{*}{pAcc$\uparrow$}& \multirow{2}{*}{cAcc$\uparrow$}& \multirow{2}{*}{FID$\downarrow$}\\ 
NCE&& Loss & Mining & & & & & \\
\hline
\checkmark&$\times$&$\times$ &$\times$&45.5 &24.7&68.8&30.7&56.4\\ 
\checkmark&$\times$&\checkmark &$\times$&42.7&27.2&73.0&33.1&52.7\\ 
\checkmark&$\times$&\checkmark&\checkmark& 40.5 &27.4 &72.8&33.5&49.9\\ 
\hdashline
$\times$&\checkmark&$\times$&$\times$ & 41.2&27.8&72.8&33.4&52.0\\ 
$\times$&\checkmark&$\times$&\checkmark&  37.6& 27.3& 71.4&33.4&50.1\\ 
$\times$&\checkmark&\checkmark&$\times$&  36.2& 27.9& 73.3&34.2&49.9\\ 
\hdashline
$\times$&\checkmark&\checkmark&\checkmark&\textbf{34.4}& \textbf{29.0}&\textbf{73.5} &\textbf{35.6} &\textbf{46.4} \\
\hline

\end{tabular}
}
% \end{adjustbox}
\end{center}
    \vspace*{-.1cm}

\vspace*{-.5cm}
\caption{Quantitative results of ablation studies.}

\vspace*{-.4cm}
\label{table:ablation}
% \end{threeparttable}
\end{table}

\section{Conclusion}
In this paper, we proposed a novel method utilizing the relational knowledge formed by the heterogeneous semantics of image patches for image translation tasks. We presented the decoupled infoNCE with hard negatives (hDCE) along with the regularization of semantic relation consistency (SRC) to improve the performance. We verified our methods using various tasks, which are image translation tasks and GAN compression. The results have shown noticeable performance increase, achieving state-of-the-art scores when compared to the baseline models. Discussions about limitations and negative social impacts are in our Supplementary Materials.

\end{document}